%% file: main.tex
\documentclass{article}
\usepackage{spconf,amsmath,graphicx}

\usepackage{graphicx}
\usepackage{amssymb,amsmath,bm}
\usepackage{textcomp}
\usepackage{booktabs}
\usepackage{siunitx}
\usepackage{xspace}
\usepackage{url}
\usepackage{lipsum}  
\usepackage{tipa}
\usepackage{enumitem}
\usepackage{hyperref}

\graphicspath{{figures/}}

\input{custom_commands}

\begin{document}
\title{A STUDY ON THE IMPACT OF SELF-SUPERVISED LEARNING ON AUTOMATIC DYSARTHRIC SPEECH ASSESSMENT}

\input{meta/authors}
\maketitle

\input{sections/abstract}

\input{meta/keywords.tex}
\input{sections/introduction}
\input{sections/related_works}

\input{sections/results}

\input{sections/discussion}

\input{meta/acknowledgments}

\clearpage %
\bibliographystyle{IEEEbib}
\bibliography{mybib}

\end{document}

%% file: custom_commands.tex
\newcommand{\khz}[1]{
  \SI{#1}{\kilo\hertz}
}

%% file: meta/authors.tex
\name{Xavier F.~Cadet\thanks{Corresponding author: xfc17@ic.ac.uk} \qquad Ranya Aloufi \qquad Sara Ahmadi-Abhari \qquad Hamed Haddadi}
\address{Imperial College London}

%% file: sections/abstract.tex
\begin{abstract}
    
Automating dysarthria assessments offers the opportunity to develop practical, low-cost tools that address the current limitations of manual and subjective assessments.
Nonetheless, the small size of most dysarthria datasets makes it challenging to develop automated assessment.
Recent research showed that speech representations from models pre-trained on large unlabelled data can enhance Automatic Speech Recognition (ASR) performance for dysarthric speech.
We are the first to evaluate the representations from pre-trained state-of-the-art Self-Supervised models across three downstream tasks on dysarthric speech: disease classification, word recognition and intelligibility classification, and under three noise scenarios on the UA-Speech dataset.
We show that HuBERT is the most versatile feature extractor across dysarthria classification, word recognition, and intelligibility classification, achieving respectively $+24.7\%, +61\%, \text{and} +7.2\%$ accuracy compared to classical acoustic features.

\end{abstract}

%% file: meta/keywords.tex
\noindent\textbf{Index Terms}: 
dysarthric speech, speech recognition, self-supervised learning

%% file: sections/introduction.tex
\section{Introduction}
Dysarthria is caused by a lack of articulatory control and muscle weakness, which affect speech rate, dynamic amplitudes and pitches, and how the spoken word is pronounced.
All of these contribute to unintelligible speech and difficulty understanding due to the inaccurate articulation of phonemes and abnormal speech patterns~\cite{mendoza2021effect-COP}.
Dysarthria classification has become increasingly important in diagnosing the disorder, determining the best treatment options, and conducting speech therapy sessions as needed~\cite{%
yue22_interspeech-COP}. Nonetheless, obtaining dysarthric speech samples is usually challenging, as most datasets contain a small number of speakers. Furthermore, there is limited research on how well these assessments can provide specific performance insights for individual patients.

Self-Supervised learning (SSL) in speech processing enables learning from large, unlabeled datasets, enhancing the understanding of diverse speech patterns~\cite{hsu2021hubert-COP}.
While recent research has shown that SSL approaches can outperform supervised ones in dysarthric Automatic Speech Recognition~\cite{qian2023survey,violeta22_interspeech-COP}, it has been not evaluated for other dysarthric assessments: disease, word and intelligibility classification and under various noise patterns.
This research gap underscores the need to examine SSL approaches in different dysarthric assessments and environment conditions. Understanding the types of impairments and their patterns better can aid in developing better tools for identifying disorders and their traits.

Our~\textbf{main contribution} is the evaluation of representations extracted by Self-Supervised models trained on large scale healthy speech under three noise scenarios, across three classification tasks.
We propose a tool to empirically evaluate different representations (e.g., acoustics and self-supervised methods) and classifiers (e.g., Logistic Regression (LR) and Multi-Layer Perceptron (MLP)), on binary and multi-class tasks (e.g., disease, word, and intelligibility classification).
To simulate real-world recording collection scenarios, we experiment under three settings: default, noise addition, and noise reduction.
Our tool can provide insight on which features can facilitate the identification of disorders and their characteristics.
For instance, for the severity classification under these three scenarios, we showed that SSL feature extractors pretrained on healthy speech can be applied on dysarthric speech and still provide good performance.

%% file: sections/related_works.tex
\section{Dysarthria Automatic Assessments}
Dysarthria intelligibility assessment is typically performed in two stages~\cite{hall2022investigation-COP}.
The training stage involves building a computational model based on patients’ speech samples and their respective speech intelligibility classes. After training the model, one can identify classes of speakers with unknown intelligibility levels, by comparing their acoustic features with those used during training. Reference-free intelligibility assessment approaches focus on developing classification models without any prior understanding of healthy speech. Instead, they focus on extracting acoustic features believed to be highly correlated with intelligibility ~\cite{falk2012characterization-COP}.
Meanwhile, reference-based approaches utilize healthy speech data (e.g., ASR-based approaches) to determine the characteristics of intelligible speech and use them as a basis for estimating the level of intelligibility~\cite{%
qian2023survey, tripathi2020improved-COP}.
Such approaches exploit the fact that ASR systems trained only on healthy speech perform poorly on dysarthric speech and that the performance of ASR systems deteriorates with the severity of dysarthric speech.

%% file: sections/results.tex
\section{%
  Methodology}
\input{tables/benchmark}

\begin{figure}[!t]
      \centering  \includegraphics[width=0.5\textwidth]{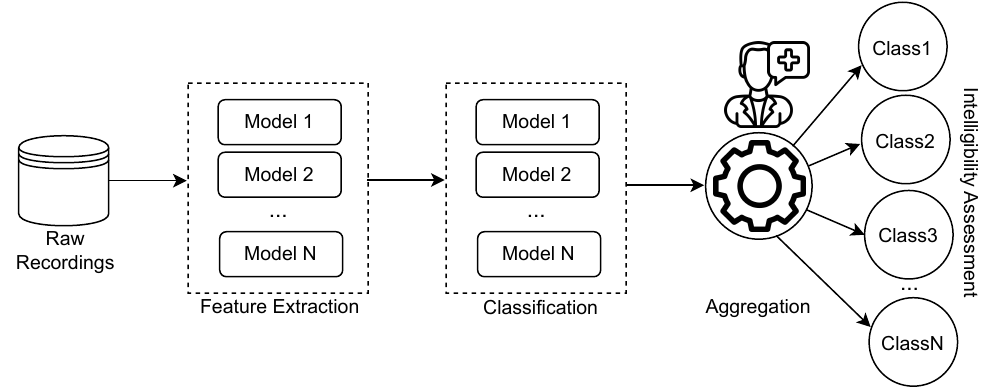}
      \caption{The proposed tool overview.
      }
      \label{fig:overall}
      \vspace{-5mm}
\end{figure}

\subsection{Overview}
\label{sec:overview}
The quality of recordings, representations used, and classification algorithm can influence the effectiveness of automated dysarthria assessments. Therefore, we aim to develop an interpretable tool that facilitates understanding the outputs of such assessments.
An overview of the proposed tool is provided in Figure~\ref{fig:overall}, which can be easily adapted to extract various features, followed by multiple classification algorithms.
Then, our tool aggregates the results per patient to verify the assessment results' reliability.
Aggregation outputs could be interpreted as intelligibility classes, such as low, mid, and high levels, and could provide clinicians with an interpretable classification of the speaker's intelligibility.
\subsection{Experimental Setting}
\textbf{Dataset.}
The UA-Speech~\cite{kim2008dysarthric-COP} dataset contains recordings from 13 healthy control speakers and 15 dysarthric speakers.
The vocabulary includes 455 distinct words with ten digits, 26 radio alphabets, 19 computer commands, 100 common words, and 300 uncommon words.
Speakers are divided into four different categories based on the severity of the condition, namely high (H), mid (M), low (L), and very low (VL).

\textbf{Hand-crafted Features.}
We extracted acoustic measure of articulation, voice and prosody using PRAAT. ~\cite{%
      joshy2022automated-COP}.
Examples include the mean harmonic-to-noise ratio (HNR), the fraction of locally unvoiced frames, the number of voice breaks, degree of voice breaks, the mean and standard deviation of pitch, jitter, and shimmer, and cepstral peak prominence (CPP)~\cite{tran22c_interspeech-COP}.

\textbf{Self-supervised Features.}
We used pre-trained Self-Supervised Feature extractors provided by the SUPERB benchmark 
\footnote{\href{https://github.com/s3prl/s3prl}{https://github.com/s3prl/s3prl}}: wav2vec2, Modified CPC~\cite{9054548-modified-cpc}, and HuBERT.

\textbf{Dysarthria Classification.}
We treated the Dysarthria classification as a binary classification task; a participant is either in the control group, class 0, or the dysarthria group, class 1.

\textbf{Word Classification.}
Dysarthria patients are more likely to be able to utter isolated words rather than continuous sentences~\cite{qian2023survey}.
Isolated word recognition converts the input speech command into the corresponding text format~\cite{fu2020application-COP}.
Keyword spotting involves detecting specific words or phrases within longer spoken sentences or utterances~\cite{chen2014small-COP}.
We designed this task as a multi-classification task with $155$ individual words.
We considered only words not identified as uncommon in the original dataset.

\textbf{Intelligibility Classification.}
Dysarthria can vary in severity, leading to speeches of different degrees of intelligibility~\cite{kim2011acoustic-COP}.
We considered five classes: the four directly available from the UA-Speech dataset, and one to represent control speakers.

\textbf{Classification Models \& Evaluation Metric.}
We compared the performance of Logistic Regression (LR) and Multi-Layer Perceptron (MLP) classifiers. We reported results in terms of accuracy using a Leave-One-Speaker-Out (LOSO) approach.
We evaluated the performance at the recording level.
For instance, an audio sample is classified as dysarthric or not.

\begin{figure*}[!t]
      \centering  \includegraphics[width=\textwidth]{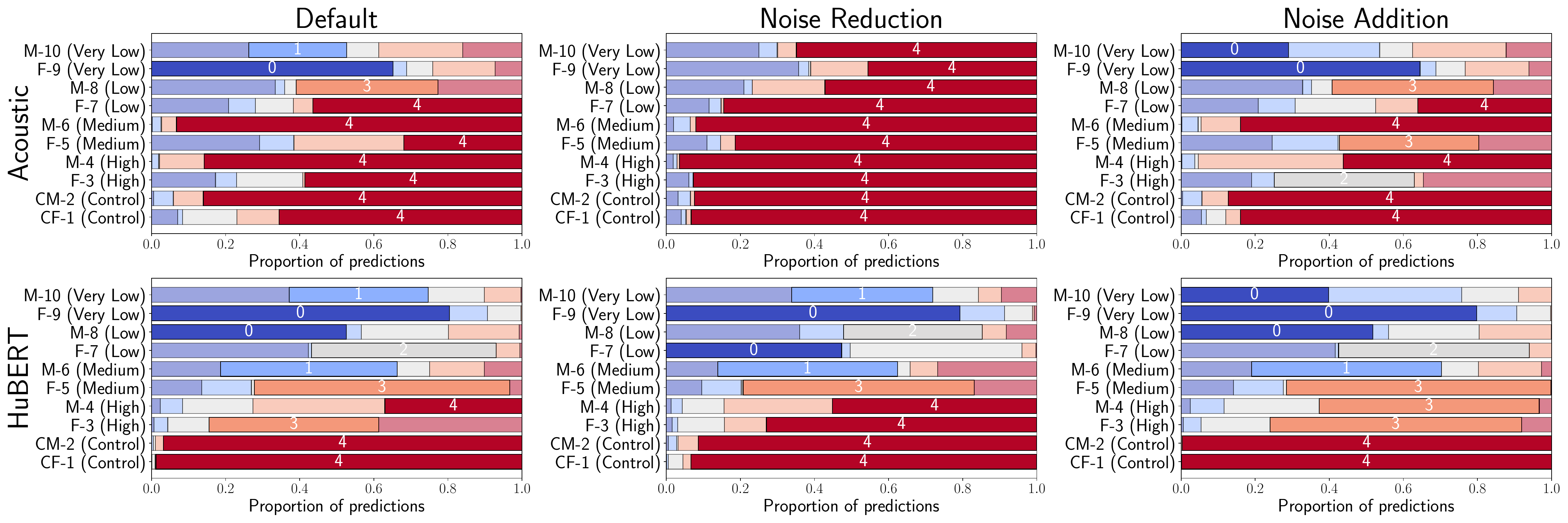}
      \caption{Patient-level predicted intelligibility:
        The top and bottom row show the predictions using respectively the acoustic features and the HuBERT features.
        The predictions are reported from left to right based on the environment: Default, Noise Reduction, and Noise Addition datasets.
        Each intelligibility class is gradient-color coded, from very low intelligibility in blue  on the left to control level in red on the right.
        For each patient the section that stands out indicates the majority predicted intelligibility class, along with its label (Very Low: 0, Low: 1, Medium: 2, High: 3, Control: 4).
        While the performance based on HuBERT %
        features is higher than acoustic, for a given speaker, there are major mis-classifications at the recording level.
      }
      \label{fig:speaker_intelligibility}
      \vspace{-3mm}
\end{figure*}

\input{sections/pipeline}

%% file: tables/benchmark.tex
\begin{table*}[t]
    \input{tables/captions/benchmark.tex}
    \vskip 0.1in
    \begin{center}
        \begin{small}
            \begin{sc}
                \begin{tabular}{cccccccc}
                    \toprule
                               &     & \multicolumn{2}{c}{Default} & \multicolumn{2}{c}{Noise Reduction} & \multicolumn{2}{c}{Noise Addition}                      \\
                   Task  & Extractor & \multicolumn{2}{c}{Accuracy $\uparrow$} & \multicolumn{2}{c}{Accuracy $\uparrow$} & \multicolumn{2}{c}{Accuracy $\uparrow$}                      \\
                                    &              & LR                          & MLP                                 & LR                                 & MLP  & LR   & MLP  \\
                    \midrule
                    \midrule
                    disease         & acoustic     & 69.3 (22.6) & 65.8 (24.6) & 59.4 (23.8) & 54.8 (33.7) & 79.4 (17.9) & 76.6 (21.0) \\
                    \midrule
                    disease         & wav2vec2     & 94.2 (8.0) & 94.1 (8.5) & 82.8 (16.4) & 81.8 (17.9) & 99.8 (0.2)  & 99.7 (0.4)  \\
                    disease         & hubert       & 94.0 (8.0) & 93.4 (9.8) & 84.8 (15.9) & 85.5 (15.1) & 99.7 (0.5) & 99.4 (1.3) \\
                    disease         & modified cpc & 94.9 (7.3) & 95.8 (6.7) & 82.0 (19.4) & 84.1 (15.6) & 99.7 (0.5) & 99.6 (0.9) \\
                    \midrule
                    \midrule
                    words           & acoustic     & 9.6 (5.3) & 9.6 (5.7) & 8.9 (5.2) &  11.3 (6.4) & 9.2 (5.0) & 9.6 (5.7) \\
                    \midrule
                    words           & wav2vec2     & 56.5 (27.4) & 56.5 (27.2) & 43.9 (23.8) & 44.4 (24.0) & 54.0 (26.5) & 54.5 (26.3) \\
                    words           & hubert       & 70.6 (29.1) & 69.3 (29.0) & 56.6 (27.7) &  55.8 (27.4) & 70.2 (29.0) & 69.1 (29.1) \\
                    words           & modified cpc & 53.8 (29.4) & 57.3 (29.1) & 46.1 (25.8) & 48.1 (26.5) & 54.7 (30.2) & 57.1 (29.5) \\
                    \midrule
                    \midrule
                    intelligibility & acoustic     & 45.5 (34.8) & 45.1 (32.7)	& 46.3 (41.8)	 & 39.3 (36.8) & 53.5 (36.4)  & 53.2 (34.9) \\
                    \midrule
                    intelligibility & wav2vec2     & 63.4 (37.5) &63.2 (37.6)	 & 54.3 (35.5)	 & 54.6 (36.0) & 67.6 (38.5)  & 67.4 (38.1)  \\
                    intelligibility & hubert       & 61.6 (39.1)	 & 62.6 (37.9)	 & 55.4 (36.2) & 55.5 (35.9) & 66.2 (40.2)  & 66.9 (39.5)  \\
                    intelligibility & modified cpc & 59.6 (41.1)	 & 62.4 (39.9)	 & 52.3 (35.0) & 54.1 (34.6)	 &  62.9 (41.9)  & 64.9 (41.0) \\
                    \bottomrule
                \end{tabular}
            \end{sc}
        \end{small}
    \end{center}
    \vspace{-5mm}
\end{table*}

%% file: tables/captions/benchmark.tex
\caption{Task Performance:
    We report the mean and (standard deviation) of the accuracy.
    Results for disease and word classification are derived from individual recordings, whereas intelligibility task results are based on speaker-level performance.
}
\label{table:performance}

%% file: sections/pipeline.tex
\section{Results}
\textbf{[Default Setting]}~\textbf{Q1.}~\emph{How reliable are classifiers when confronted with unknown speakers?}

\textbf{Preliminaries.}
Healthcare professionals can benefit from classification tasks to better understand and manage speech difficulties associated with dysarthria.
While the literature references existing attempts at automated assessment that attain high levels of accuracy, their generalizability might be biased due to their evaluation methodology~\cite{%
hall2022investigation-COP}.
Huang et al., ~\cite{huang2021review-COP} show that most studies on automated dysarthria assessment achieve high accuracy, ranging from 75.1\% to 93.97\%.
The models are often trained and evaluated on the same speakers or only one unseen speaker for each target class~\cite{al2021classification-COP}, potentially leading to biased results due to the model's ability to identify the speaker rather than dysarthria-related information. 
\newline
\textbf{Setup.}
We assessed the classifiers' performance against speakers absent during training.
The UA-Speech dataset's raw recordings are categorized by patient ID to ensure no overlap between the training and testing sets. Thus, all the recordings from a speaker are either in the training or test set.
We used hand-crafted features, including acoustic features, and various representations obtained from %
self-supervised models like HuBERT, wav2vec2, and Modified CPC for feature extraction.
\newline
\textbf{Results.}
Models trained with self-supervised representations outperformed models trained on acoustic features for all dysarthria assessment tasks (Table ~\ref{table:performance}).
The results are obtained without fine-tuning the self-supervised feature extractors, making them a promising direction for automated dysarthria assessment.
The HuBERT and wav2vec2 representations demonstrated a word recognition accuracy ranging from 56.5\% to 70.6\%, surpassing the acoustic models that achieved only 9.6\% accuracy. Similarly, these models showed accuracies (around 63\%), significantly higher than acoustic models (45.5\% for LR and 45.1\% for MLP) in the intelligibility task. 

To better understand the reliability of the assessment at the patient level, we propose a tool in Section~\ref{sec:overview}, which allows for a detailed analysis of the predictions (Figure \ref{fig:speaker_intelligibility}).
One can adapt the proposed tool to other feature extractors and classification models.
This tool can provide a more interpretable assessment per patient, and lead to personalized treatments.
\newline
\newline
\textbf{[Noise Reduction]}~\textbf{Q2.}~\emph{What impact does enhancing the recordings have over the different tasks?}

\textbf{Preliminaries.}
We considered a scenario under which we enhance the Default dataset.
We considered speech restoration, a process that aims at restoring degraded speech signals to their original quality~\cite{godsill2013digital-COP}.
For instance, speech is typically surrounded by background noise, blurred by reverberation in the room, or recorded with low-quality equipment.
Ambient noise from clinical clicks or other artifacts may be present in dysarthric recordings. %

\textbf{Setup.}
Our objective was to enhance the recordings by applying one of the speech enhancement approaches and evaluate the models' performance in such scenarios.
We generate a new version of the dataset after applying `VoiceFixer'~\cite{liu2021voicefixer-COP}, a method that attempts to remove multiple distortions simultaneously.
We apply resampling to \khz{16} before extracting the representations to ensure that the sampling rates match between the input signal and the feature extractor.
\newline
\textbf{Results.}
Across all tasks, feature extractors and classifiers, the performance decreased (Table \ref{table:performance}).
Nonetheless, self-supervised models still outperformed acoustic models (52.3\% to 55.5\% versus 46.3\% for LR and 39.3\% for MLP) in intelligibility task for instance.
We looked deeper into patient-level variations using our proposed visualization tool to display the model's predictions for the speaker intelligibility task.
Acoustic features-based models predicted most participants as the control intelligibility group (Figure \ref{fig:speaker_intelligibility}). 
Furthermore, the enhancement tool led to partial speech segment removal in some recordings.
As such, the systematic use enhancement tools requires additional care if used as a preprocessing step in automated assessment pipelines.
\newline
\newline
\textbf{[Noise Addition]}~\textbf{Q3.}~\emph{Are models trained on the Default dataset biased by patterns related to the noise in recordings?}
\newline
\textbf{Preliminaries.}
Some recordings in the default settings of the dataset have different noise levels and clicking sounds.
Thus, one could argue that the good performance observed on the disease classification task is due to patterns associated with external factors from the recording rather than speech-related information.
To confirm whether the feature extractors and models can leverage information specific to classes, we conduct audio sample mixing to exacerbate such an effect.
\newline
\textbf{Setup.}
We generated a new version of the dataset to determine whether the models perform well because they leverage specific noise patterns or if they extract information from the speech itself.
First, we obtained a single background noise sample from the WHAM dataset~\cite{wichern2019wham-COP}.
Then, we mixed every audio recording from \textit{control patients} with that noise pattern. Under such a scenario, feature extractors and models able to extract that singular noise pattern would achieve higher performance.
\newline
\textbf{Results.}
All combinations of features extractors and models achieved higher accuracy on the classification task (Table \ref{table:performance}).
With noise addition, self-supervised models demonstrated great performance, with accuracies close to 99.7\% in disease detection (binary task) and maintaining accuracies above 62.9\%, compared to the acoustic models' 53.5\% for LR and 53.2\% for MLP in intelligibility (multi-label task). For word recognition, HuBERT maintained high performance (around 70\%), slightly higher than wav2vec2 and Modified CPC.
Therefore, all feature extractors and models could leverage such a pattern when enforcing a singular pattern across control patients.

%% file: sections/discussion.tex
\section{Conclusions and Future Work} 
\textbf{Self-Supervised Learning in Dysarthria.}
Self-supervised representations, such as HuBERT, wav2vec2, and Modified CPC, demonstrated higher performance in all dysarthria evaluation tasks (i.e., disease detection, word recognition, and intelligibility). As such, feature extractor trained on large scale healthy speech datasets can be leveraged for smaller dataset with dysarhtic speech. 
\newline
\textbf{Patient-level Inspection.}
We proposed a tool to inspect the predictions at a patient level.
Given a reliable classifier, one can inspect the different predictions for a given patient and determine whether the predictions indicate a mix of classes or if a class is overwhelmingly represented.
\newline
\textbf{Classes Imbalance.}
For the Intelligibility Severity assessment, there is a major class imbalance concerning the number of recordings, \textit{i.e.,} $8.3\%$ (VL), $14\%$ (L), $13.9\%$ (M), $63.8\%$ (H + C) or $19.6\%$ (H) and $44.2\%$ (C).
Furthermore, the intelligibility labels are coarse.
For instance, patients intelligibility score between $26\%$ and $49\%$ are grouped together.
The imbalance and coarse labels make it challenging to obtain fine-grained predictions.
\newline
\textbf{Limitations and Future Work.}
Future work would benefit from exploring Whisper representations as well, and fine-tuning the feature extractors on dysarthric data.
Nonetheless, despite not requiring labels, Self-Supervised pretraining directly on the imbalanced dysarthric data could  lead to representations that benefit the high/control intelligibility representations. As such, future works would benefit from leveraging methods to tackle class imbalance.
In addition, the current study focused on dysarthria, but it is unclear how these SSL models perform with other speech impairments. The generalizability of these models across a broader range of speech disorders remains an area to explore, and further investigation on other datasets would be beneficial.

%% file: meta/acknowledgments.tex
\section{Acknowledgements}
We thank Sandra Siby for her invaluable suggestions that improved the text.
Xavier F. Cadet is supported by UK Research and Innovation (UKRI Centre for Doctoral Training in AI for Healthcare grant number EP/S023283/1).
Hamed Haddadi is supported by the EPSRC Open Plus Fellowship (EP/W005271/1: Securing the Next Billion Consumer Devices on the Edge).
For open access, the authors have applied a Creative Commons Attribution (CC BY) license to any Author Accepted Manuscript version arising.